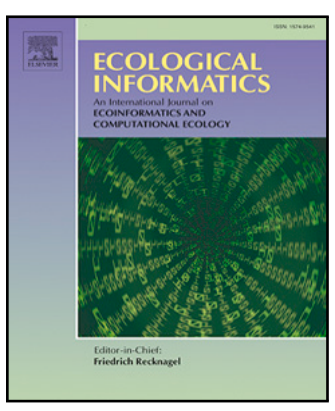

# CoralSCOP-LAT: Labeling and analyzing tool for coral reef images with dense semantic mask

Yuk Kwan Wong [a], Ziqiang Zheng [a,*], Mingzhe Zhang [a], David J. Suggett [b,c], Sai-Kit Yeung [a]

[a] Hong Kong University of Science and Technology, Hong Kong
[b] King Abdullah University of Science and Technology, Saudi Arabia
[c] University of Technology Sydney, Australia

ABSTRACT

Coral reef imagery offers critical data for monitoring ecosystem health, in particular as the ease of image datasets continues to rapidly expand. Whilst semi-automated analytical platforms for reef imagery are becoming more available, the dominant approaches face fundamental limitations. To address these challenges, we propose `CoralSCOP-LAT`, a coral reef image analysis and labeling tool that automatically segments and analyzes coral regions. By leveraging advanced machine learning models tailored for coral reef segmentation, CoralSCOP-LAT enables users to generate dense segmentation masks with minimal manual effort, significantly enhancing both the labeling efficiency and precision of coral reef analysis. Our extensive evaluations demonstrate that `CoralSCOP-LAT` surpasses existing coral reef analysis tools in terms of *time efficiency*, *accuracy*, *precision*, and *flexibility*. `CoralSCOP-LAT`, therefore, not only accelerates the coral reef annotation process but also assists users in obtaining high-quality coral reef segmentation and analysis outcomes.

## 1. Introduction

Coral reefs (Edwards et al., 2017; Cinner et al., 2016; Sandin et al., 2022; Levy et al., 2022; Neal et al., 2017; Boström-Einarsson et al., 2020) are among the most diverse and valuable ecosystems on our blue planet, creating habitats that harbor an estimated 32% of all named marine species (Cinner et al., 2016; Zheng et al., 2024b,a). Coral reef landscapes (Suggett et al., 2022) are rich in symbiotic relationships among marine species (Baker, 2003). Maintaining the health (Knowlton et al., 2021) and diversity (Boström-Einarsson et al., 2020; Levy et al., 2022) of coral reefs is crucial for the livelihoods and well-being of millions of people worldwide. Furthermore, coral reefs are a focal point for extensive research in coral ecology (Kar et al., 2022; Eddy et al., 2021; Yuan et al., 2024) and investigations into the mechanisms of coral health assessments (Helgoe et al., 2024; Harrison, 2024; Reimer et al., 2024). However, coral reefs remain in a state of decline from global and local stressors that particularly drive increased frequency and severity of bleaching and mortality events across the world (Chung et al., 2024; Lyu et al., 2022; Datta et al., 2024). These events not only degrade the structural integrity of coral ecosystems but also disrupt the broader and interconnected marine environments. Rapid, responsive, and comprehensive monitoring and surveying of coral reefs (Sandin et al., 2022; Ziqiang et al., 2023; Hughes et al., 2018; González-Rivero et al., 2019) is thus imperative for conservation efforts and sustainable management of marine resources.

To accelerate the coral reef analysis performance, various algorithms and analysis platforms (Gomes-Pereira et al., 2016; Zheng et al., 2023; Beijbom et al., 2015; Chen et al., 2021b) have been proposed, which can fall into two main categories: sparse point-based analysis (Kohler and Gill, 2006; Mahmood et al., 2016; Modasshir and Rekleitis, 2020; Langenkämper et al., 2017; González-Rivero et al., 2019) and dense segmentation-based analysis (Beijbom et al., 2016; King et al., 2019; Pavoni et al., 2019; Li et al., 2024; Zhong et al., 2023a).

### 1.1. Sparse point-based approaches

Sparse point-based approaches scatter random points/pixels across images to classify features based on underlying semantic categories (*e.g.,* coral, algae, rubble). While these methods are straightforward, they often lead to over- or under-estimation of environment indices, because the results are highly dependent on sampling density. Point annotations often fail to capture small coral formations and irregular boundaries. Further approaches, such as CoralNet (Beijbom et al., 2015;

* Corresponding author.
*E-mail addresses:* ykwongaq@connect.ust.hk (Y.K. Wong), zzhengaw@connect.ust.hk (Z. Zheng), mzhangdd@connect.ust.hk (M. Zhang), david.suggett@kaust.edu.sa (D.J. Suggett), saikit@ust.hk (S.-K. Yeung).

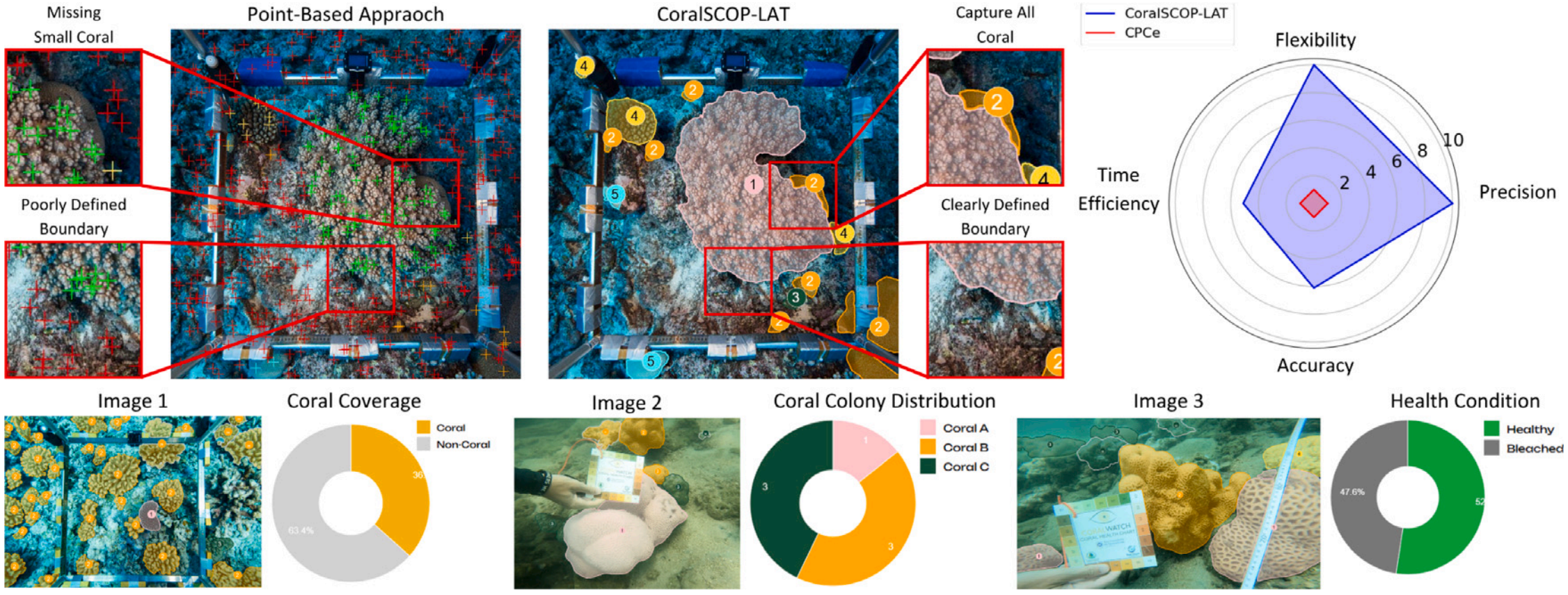

**Fig. 1.** We propose CoralSCOP-LAT, a tool for migrating coral reef analysis and labeling from sparse point-based analysis to dense mask-based analysis, which surpasses the former approach significantly in terms of *precision*, *accuracy*, *time efficiency*, and *flexibility*. Moreover, CoralSCOP-LAT supports in-depth analysis of coral reef imagery by providing the users with statistical reports. Quadrat images are credited under Under The Pole.

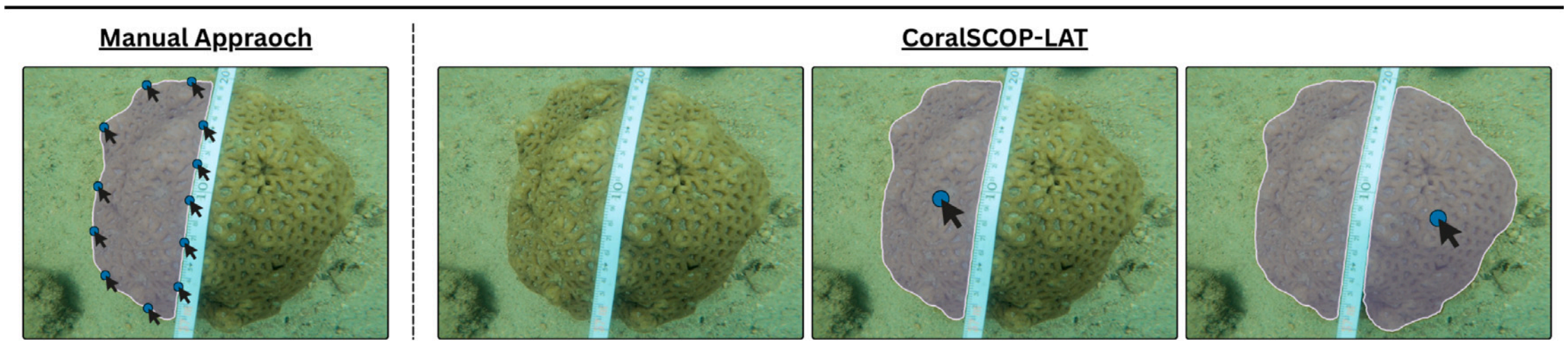

**Fig. 2.** Traditionally, creating a dense segmentation mask requires users to manually outline its boundaries. In contrast, CoralSOCP-LAT, which is integrated with SAM, enables users to generate a mask with minimal effort, requiring only a single mouse click.

Chen et al., 2021a) and ReefCloud (2025), incorporate automated coral classification to improve efficiency by assigning predefined categories based on the cropped image patches centered around sampled points. However, these approaches face challenges in complex scenarios where multiple coral instances are present within the same cropped region. Furthermore, they do not address the core limitation of sparse point-based approaches, where fine-grained and detailed features are neglected.

### 1.2. Dense segmentation-based approaches

An alternative paradigm in coral reef analysis involves pixel-wise consideration of image data, which significantly enhances precision. Sparse-to-dense conversion techniques are employed to transform sparse point annotations into dense segmentation masks. Approaches such as CoralSeg (Alonso et al., 2019) and PLAS (Raine et al., 2022) utilize Superpixels (Bergh et al., 2012) to achieve this conversion. However, these methods are prone to errors under challenging conditions, such as blurry imagery or color distortions, where boundaries are not clearly defined.

The Segment Anything Model (SAM) (Kirillov et al., 2023) has recently gained attention for segmenting complex areas. SAM accepts point or pixel inputs (also referred to as point prompts) and generates dense segmentation masks by inferring the regions associated with the point input, as illustrated in Fig. 2. While SAM demonstrates a strong capability to delineate intricate coral boundaries effectively, it depends on human-provided sparse points as input.

In addition to sparse-to-dense methods, supervised learning offers fully automated solutions for annotating irregular objects (Payal et al., 2024; Samant et al., 2023). In coral analysis, dense segmentation algorithms (King et al., 2019; Zhong et al., 2023b; Zhang et al., 2024; Li et al., 2024) are trained on annotated coral reef datasets (Beijbom et al., 2016; Edwards et al., 2017; Ziqiang et al., 2023) containing dense segmentation masks. Although these algorithms eliminate manual input, they are constrained to a predefined set of categories in the training dataset. Expanding the datasets is also labor-intensive and thus not scalable, limiting the generalization of these models to new or unseen reef sites.

CoralSCOP (Zheng et al., 2024c) addresses these limitations by introducing the first foundation model for coral reef segmentation. This model automatically segments coral reefs regardless of coral species. Moreover, it maintains high accuracy even when applied to “unseen” coral reef images, where the data distribution deviates from its training set, demonstrating its strong zero-shot ability. With its robust generalization across diverse sites, CoralSCOP represents a significant advancement, offering a foundational framework for developing efficient, general-purpose coral reef analysis tools (see Fig. 1).

### 1.3. Existing coral analytic tools

Leveraging the aforementioned methodologies, several tools have been developed for coral reef labeling and analysis (Beijbom et al., 2016; Edwards et al., 2017; Ziqiang et al., 2023). For example, Coral Point Count with Excel Extension (CPCe) (Kohler and Gill, 2006) and PhotoQuad (Trygonis and Sini, 2012) provide sparse point-based methodologies with visualization interfaces, incorporating interactive features to streamline identification processes. CoralNet (Beijbom et al., 2015; Chen et al., 2021a) offers an online platform that allows users to upload and analyze coral reef images, integrating a classification algorithm for automated coral identification. ReefCloud (2025) enhances usability by introducing user-friendly visualization and interactive functionalities, such as customizable coral/substrate codes, along with adjustable data point shapes, sizes, and colors.

**Sparse-Point-Based Approaches**

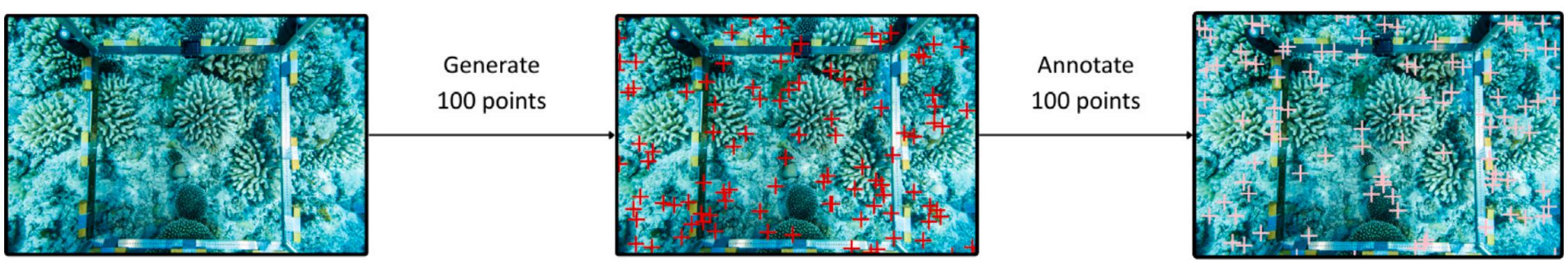

(a) Sparse point-based approaches involve generating random points on images. Users are required to annotate all the generated points.

**SAM**

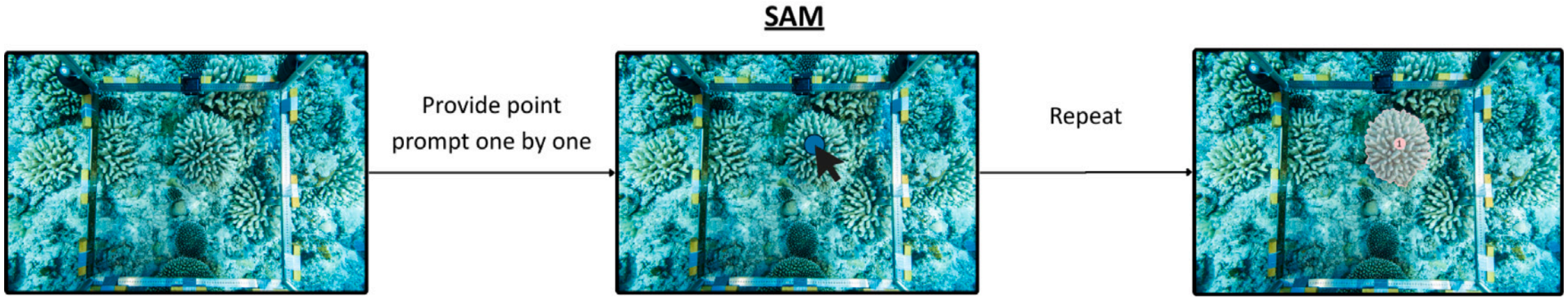

(b) SAM facilitates the generation of dense masks through a simple click. However, this approach still relies on substantial human input in complex coral reef environments.

**CoralSCOP-LAT**

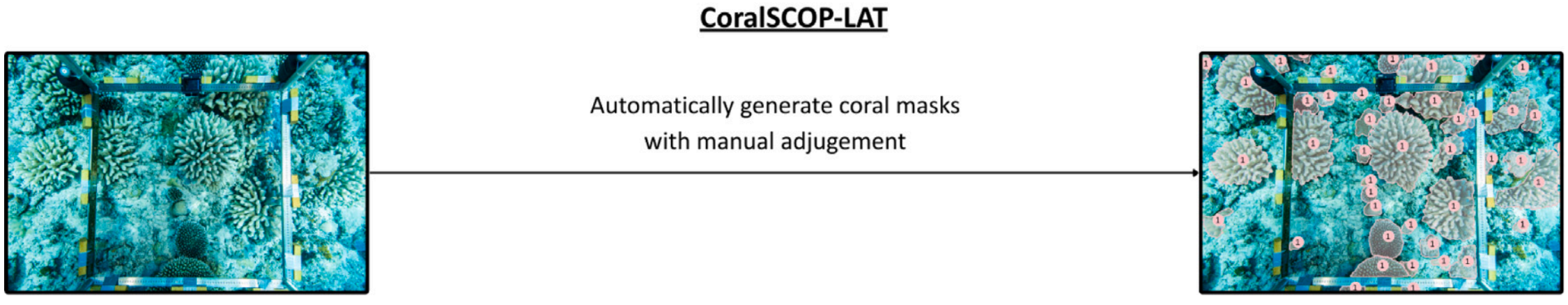

(c) `CoralSCOP-LAT` further enhances annotation efficiency by incorporating automatic coral segmentation functionalities.

**Fig. 3.** A comparison of the workflows of sparse point-based approaches, SAM-based approaches, and `CoralSCOP-LAT`. Quadrat images are credited under Under The Pole.

### *1.4. CoralSCOP-LAT*

We propose a hierarchical and user-friendly annotation tool, `CoralSCOP-LAT`, designed to enhance coral reef analysis. `CoralSCOP-LAT` offers three core features that streamline and optimize the annotation workflow:

1. **Automatic coral segmentation:** `CoralSCOP-LAT` leverages the state-of-the-art coral reef segmentation model, CoralSCOP (Zheng et al., 2024c), to automatically segment coral area within images, which can serve as the initial annotation to reduce efforts to manually detect each coral individually from scratch. As a result, analysis is significantly expedited, allowing users to use annotations directly or make minor adjustments in cases of inaccuracies. This feature distinguishes `CoralSCOP-LAT` from other annotation tools. A comparative illustration is provided in Fig. 3.
2. **Flexible analysis**. Our tool offers significant flexibility in annotation editing and coral species classification. Users can customize categories at various taxonomic levels (*e.g.,* genus or species) to meet the specific requirements of different study sites. This high level of adaptability ensures the tool's suitability for a wide range of research contexts, particularly those involving the complexity and biodiversity characteristic of coral reef ecosystems. Furthermore, unlike the traditional sparse point-based approach, where users cannot adjust the location of the generated points, `CoralSCOP-LAT` grants users full control over editing segmentation masks.
3. **Statistical analysis and visualization**. Our tool provides an intuitive user interface to generate statistical reports and visualizations. This interface enables users to interact with the machine learning models and conduct their tasks with greater efficiency and accuracy. By making the tool accessible to both technical and non-technical users, `CoralSCOP-LAT` empowers a wide range of individuals to contribute to coral analysis and conservation efforts.

With these features, `CoralSCOP-LAT` aims to streamline the analysis process for coral image datasets. Its capabilities not only boost efficiency but also support users in conducting large-scale analyses and research in marine biology, conservation, and related fields. Users can harness these advancements to conduct more comprehensive studies, ultimately contributing to a deeper understanding of coral ecosystems and their management. The contributions of this work are multifaceted and can be summarized as follows:

- `CoralSCOP-LAT` leverages the state-of-the-art coral reef foundation model to automatically detect and segment coral regions within images. It demonstrates a strong zero-shot generalization ability to unseen coral reef images.
- We have developed hierarchical functions and user-friendly interfaces, empowering users to better monitor and manage the analysis progress with ease and precision.
- `CoralSCOP-LAT` surpasses existing tools by a significant margin, achieving 7.3× more accurate and operating 5.1× faster in terms of accuracy and efficiency, respectively.

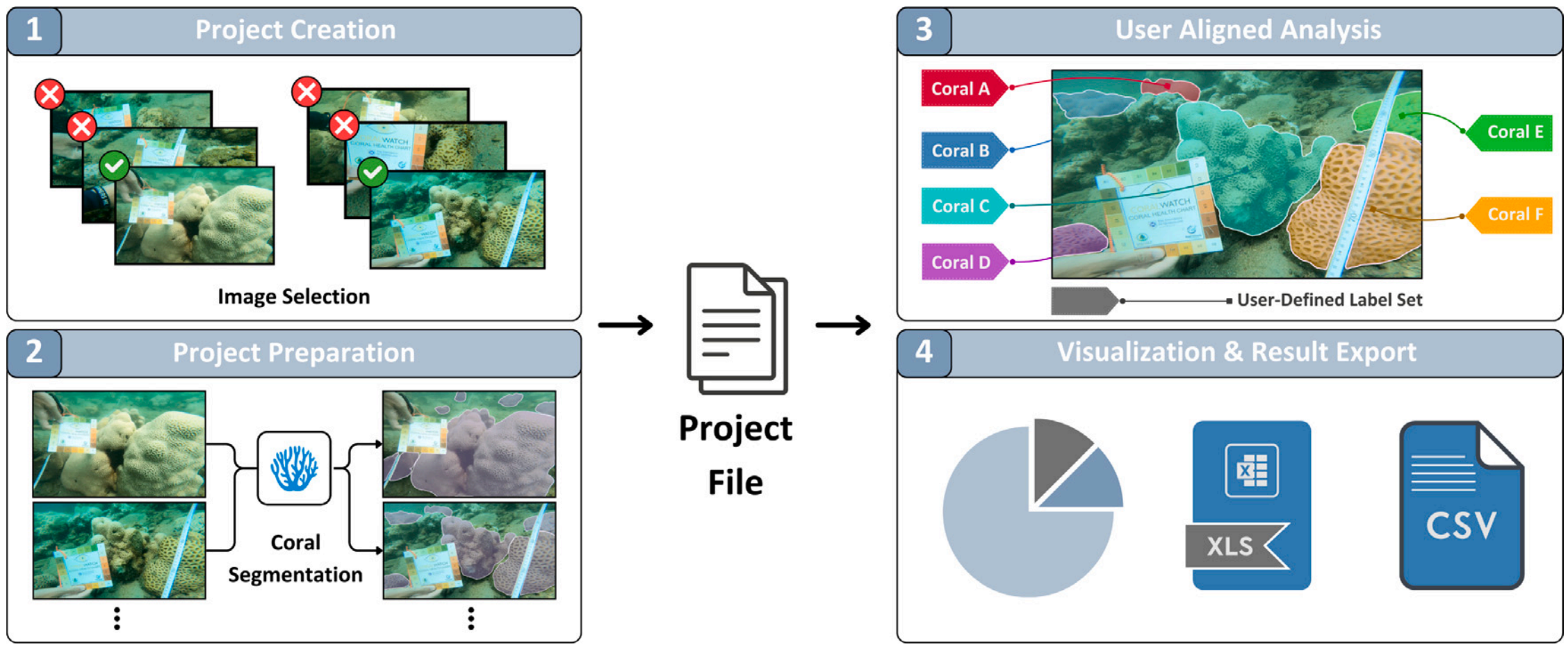

**Fig. 4.** The CoralSCOP-LAT workflow begins with users selecting target coral reef images for analysis. During the Project Preparation stage, the selected images are processed by CoralSCOP-LAT to extract image features and automatically segment coral regions. The outcomes of this preparation are saved in a project file, which can be reloaded by the user for subsequent analysis of the selected images. Additionally, the system enables the automated generation of visualizations and statistical reports based on the analyzed data.

## 2. Materials and methods

### 2.1. Workflow overview

In this section, we first provide an overview of CoralSCOP-LAT, which is depicted in Fig. 4. Users can create separate projects to manage the progress and analyze the coral reef images. We offer hierarchical functionalities for users to annotate the coral region, which is the area within an image that is occupied by corals. The main procedures of our CoralSCOP-LAT tool include:

1. **Project creation**. Users select a set of coral reef images to analyze. CoralSCOP-LAT stores the selected coral reef images into a single project for better management.
2. **Project preparation**. After project creation, CoralSCOP-LAT conducts a series of data processing on the selected images. This data processing is only required **once** for each project.
3. **User-aligned analysis**. After the preparation stage, users are able to annotate coral regions on the images. Several functionalities are provided to facilitate streamlined annotation processes.
4. **Visualization and result export**. After user-aligned analysis, users could visualize the coral reef statistics for the target image. The users could export the statistics to various formats (*e.g.*, *CSV* and *EXCEL*). Finally, the generated semantic coral reef masks can also be exported to optimize semantic coral reef segmentation algorithms.

### 2.2. Project creation

Coral reef analysis often involves processing image sets obtained through various surveying strategies. For instance, researchers may extract image frames from benthic survey videos recorded along a transect line (Urbina-Barreto et al., 2021); capture multiple top-down quadrat images of target coral reefs (Trygonis and Sini, 2012); or acquire continuous images following lawnmower pattern (Young et al., 2017). To facilitate the management of diverse image sets, CoralSCOP-LAT offers functionalities for importing and sampling image collections into one project and supports a wide range of image formats (*e.g.,* JPEG, WEBP, JPG, PNG, *etc.*). It enables users to organize data batches by batch based on the collection site.

### 2.3. Project preparation

In the project preparation phase, CoralSCOP-LAT conducts data processing on the selected images. This phase involves two key processes: automatic coral segmentation and image feature extraction.

**Automatic coral segmentation**: CoralSCOP-LAT automatically identifies and segments coral reefs within each image, reducing human efforts and time to delineate them manually. We adopted the coral reef segmentation model CoralSCOP (Zheng et al., 2024c) as it yields high-quality coral reef masks and ensures a strong generalization ability to unseen coral reef images. The user can further configure the automatic coral segmentation setting. For example, users can filter the mask for small corals by specifying the minimum area, as shown in Fig. 5.

**Image feature extraction**: CoralSCOP-LAT extracts image features from the selected images, which is a critical step enabling SAM to transform input point prompts into segmentation masks. This functionality is particularly valuable when users are dissatisfied with the automatically generated masks, as it allows for the creation of custom masks with minimal effort.

### 2.4. User-aligned analysis

After the project preparation phase, users can use the interface, as depicted in Fig. 6, to conduct analyses of coral images. This interface incorporates standard features commonly offered by existing annotation tools, including the ability to customize taxonomic categories, update labels, and visualize annotations. Additionally, it provides domain-specific functionalities, such as integrating health status awareness and generating statistical reports on environmental indices.

We demonstrate some of the basic functionalities through a workflow of a quadrant image analysis, as shown in Fig. 7. **1. Import selected image**: After the project preparation phase, users are able to import the selected images and display them on the interface, as shown in Fig. 6.

**2. Define categories**: In the second step, the users can define coral labels based on the image content. CoralSCOP-LAT provides flexible category customization, allowing users to adapt coral classifications to the specific requirements of their projects. Labels can be created according to various criteria, such as taxonomic classification, growth form, or health status, depending on the study's objectives. This adaptability

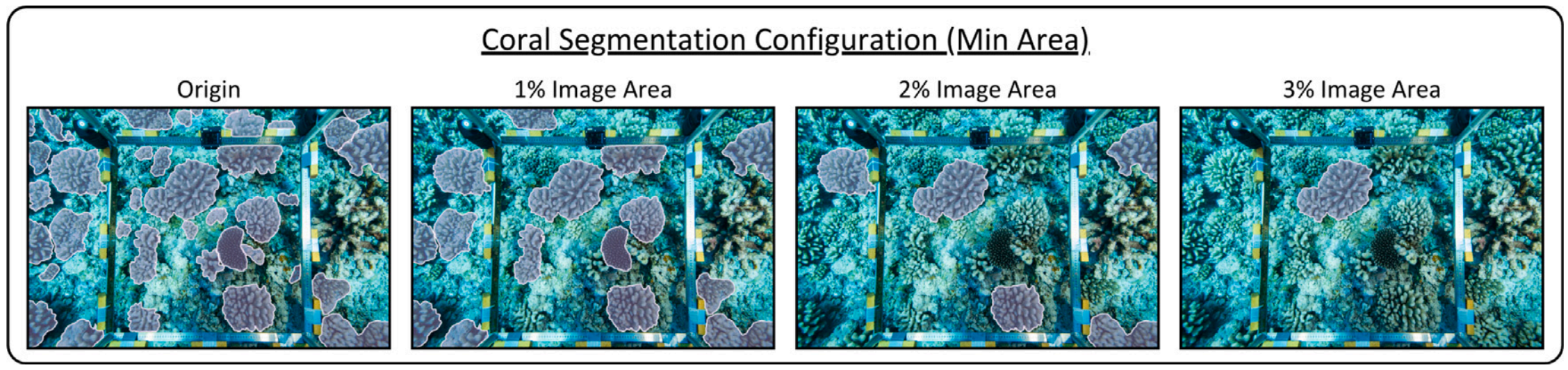

**Fig. 5.** The automatic coral segmentation algorithm can be easily configured by the user to achieve satisfactory results. For instance, the user can specify a minimum area threshold (*e.g.,* 1% of the image area) to filter out the coral regions that are too small. Quadrat images are credited under Under The Pole.

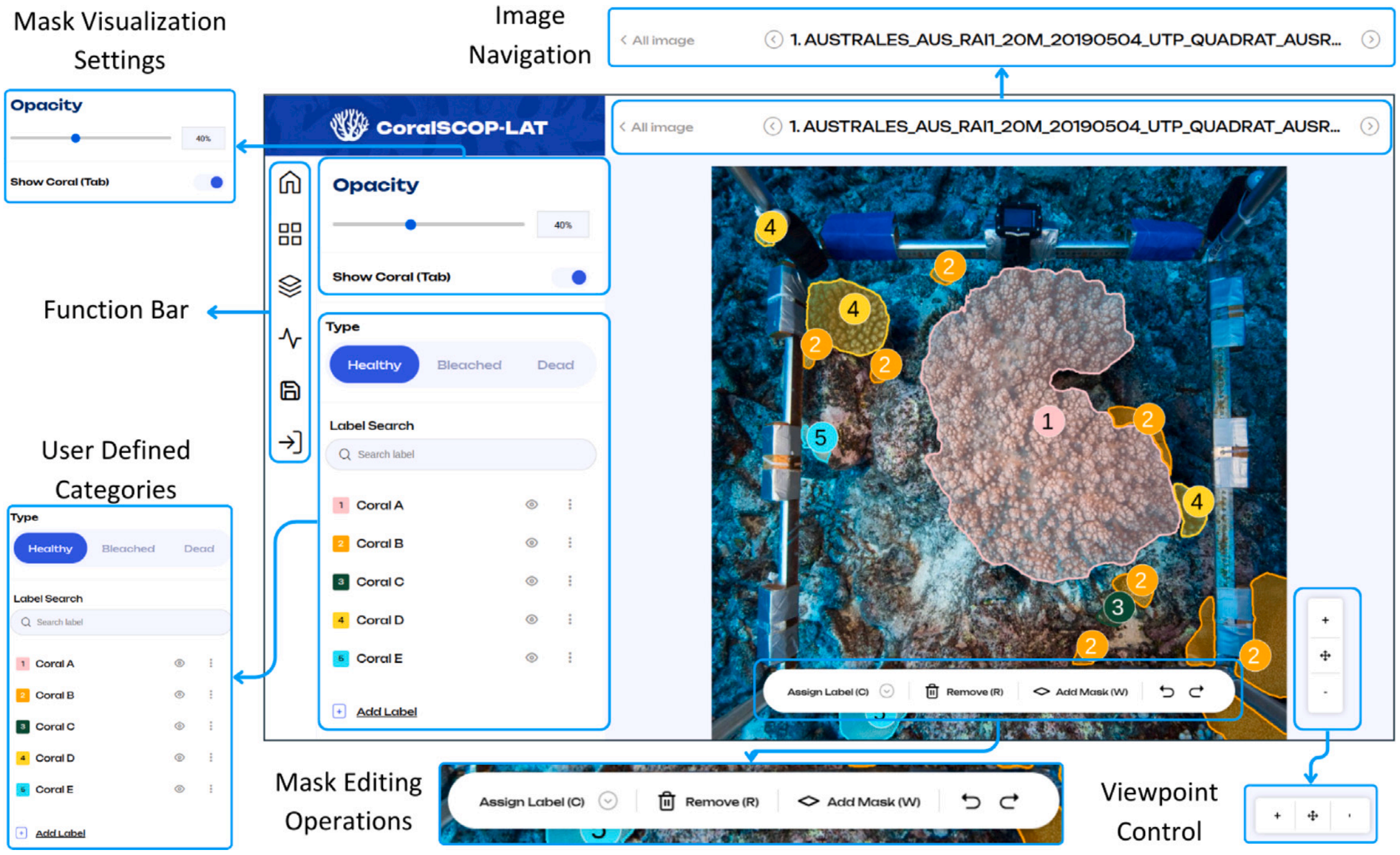

**Fig. 6.** Label Page of `CoralSCOP-LAT`, which provides a wide range of functionalities for users to identify, classify, and analyze coral reef images. Quadrat images are credited under Under The Pole.

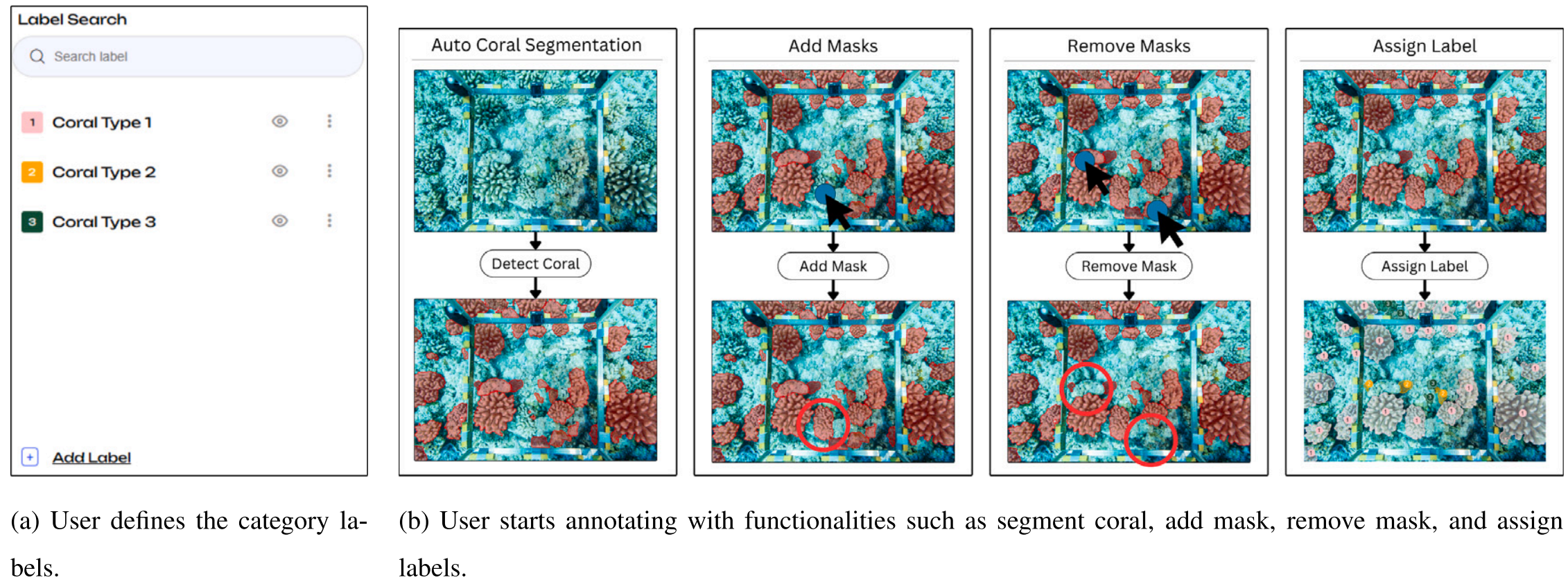

(a) User defines the category labels.

(b) User starts annotating with functionalities such as segment coral, add mask, remove mask, and assign labels.

**Fig. 7.** Overall procedural for quadrant image analysis. Quadrat images are credited under Under The Pole. (For interpretation of the references to color in this figure legend, the reader is referred to the web version of this article.)

ensures that `CoralSCOP-LAT` meets the diverse needs of different coral projects, thereby enhancing the tool's versatility and applicability across different contexts. Once the labels are defined, the categories are displayed as a list, as shown in Fig. 7(a), with each label assigned a distinct identifier and color for clear differentiation.

**3. Annotation**: After the categories are defined, the user can start annotating by using the mask editing functionalities, including automatic coral segmentation, mask addition, mask removal, and label assignment, as illustrated in Fig. 7(b). These functionalities allow users to efficiently annotate all coral instances within an image. Additionally, users can assign a health status (*Healthy*, *Bleached*, or *Dead*) to each coral instance during the annotation process. This detailed labeling facilitates in-depth analyses of coral conditions, such as calculating the bleaching percentage and mortality rate.

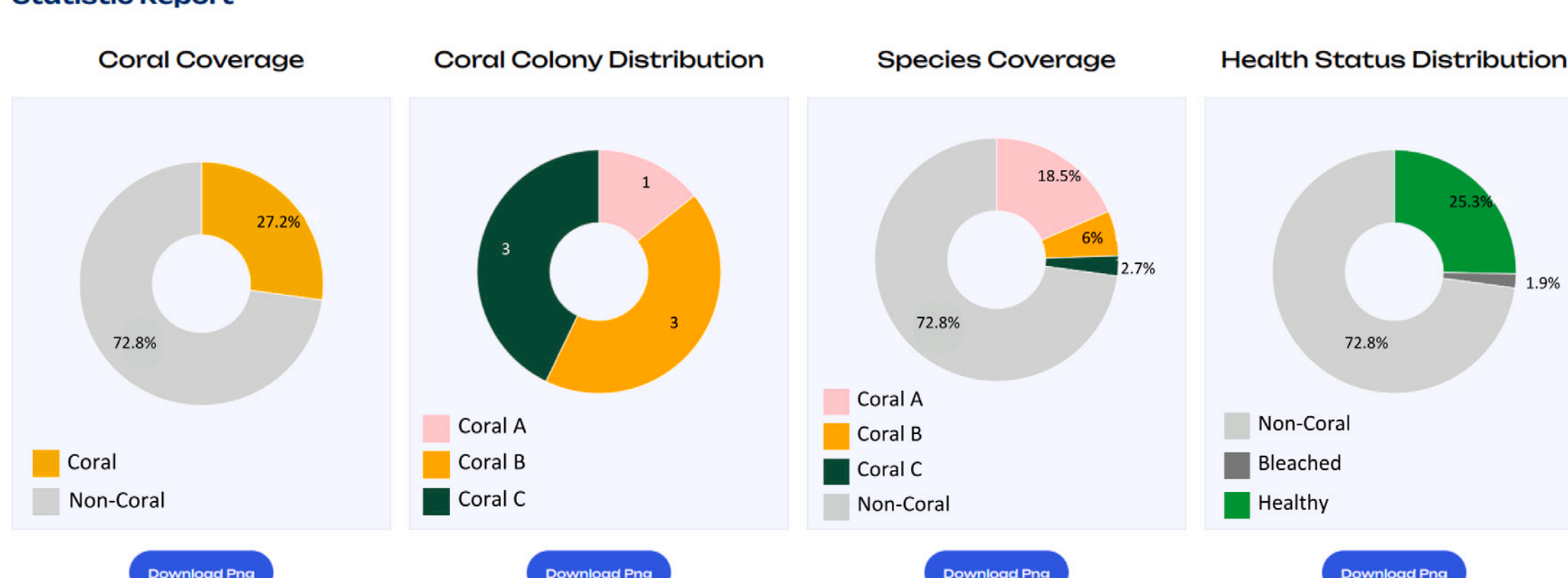

**Fig. 8.** Real-time statistic report will be generated to reveal coral condition, which includes coverage, diversity distribution, health condition summary, and so on. Users are able to download the statistical report as PNG images.

**Table 1**
Comparison of the functionalities supported across different image analysis tools. We summarize the key differences between our CoralSCOP-LAT and existing tools from comprehensive aspects.

| Functionalities supported | | CVAT | BIIGLE | PhotoQuad | CPCe | CoralNet | ReefCloud | TagLab | CoralSCOP-LAT (Ours) |
|---|---|---|---|---|---|---|---|---|---|
| Annotation (Sparse point) | Random | ✗ | ✗ | ✓ | ✓ | ✓ | ✓ | ✓ | ✗ |
| | Stratified random | ✗ | ✗ | ✓ | ✓ | ✓ | ✓ | ✓ | ✗ |
| | Uniformly distributed | ✗ | ✗ | ✓ | ✓ | ✓ | ✓ | ✓ | ✗ |
| | Manually defined | ✓ | ✓ | ✗ | ✗ | ✗ | ✗ | ✗ | ✗ |
| Annotation (Dense mask) | Manual boundary | ✓ | ✓ | ✗ | ✗ | ✗ | ✗ | ✓ | ✓ |
| | Point to mask | ✓ | ✓ | ✗ | ✗ | ✗ | ✗ | ✓ | ✓ |
| | Automatic mask generation | ✗ | ✗ | ✗ | ✗ | ✗ | ✗ | ✗ | ✓ |
| Classification | Manual | ✓ | ✓ | ✓ | ✓ | ✓ | ✓ | ✓ | ✓ |
| | Machine learning (Point) | ✗ | ✗ | ✗ | ✗ | ✓ | ✓ | ✗ | ✗ |
| | Machine learning (Mask) | ✗ | ✗ | ✗ | ✗ | ✗ | ✗ | ✗ | ✓ |
| Statistic | Coral coverage | ✗ | ✗ | ✓ | ✓ | ✓ | ✓ | ✓ | ✓ |
| | Coral species distribution | ✗ | ✗ | ✓ | ✓ | ✓ | ✓ | ✓ | ✓ |
| | Coral health statistic | ✗ | ✗ | ✗ | ✗ | ✗ | ✗ | ✗ | ✓ |
| Data export | CSV/Excel | ✓ | ✓ | ✓ | ✓ | ✓ | ✓ | ✓ | ✓ |
| | Visuazalization | ✗ | ✗ | ✗ | ✗ | ✗ | ✗ | ✓ | ✓ |
| | Statistic graphs | ✗ | ✗ | ✗ | ✗ | ✗ | ✗ | ✗ | ✓ |
| | COCO | ✗ | ✗ | ✗ | ✗ | ✗ | ✗ | ✗ | ✓ |

### 2.5. Visualization and result export

Upon completing the annotation process, CoralSCOP-LAT allows users to generate a real-time statistical report that provides insights into metrics such as coral coverage and species distribution (see Fig. 8). These reports can be downloaded as PNG images for further analysis. Users could export the statistics to various formats (*e.g.*, *CSV* and *EXCEL*). Finally, the generated semantic coral reef masks can also be exported to optimize semantic coral reef segmentation algorithms.

## 3. Results

In this section, we compare CoralSCOP-LAT with existing coral analysis tools for various features.

### 3.1. Functionalities supported

We first compare the functionalities offered by CoralSCOP-LAT with those of other well-known analysis tools in Table 1, including general-purpose image analysis tools (*e.g.*, CVAT and BIIGLE (Langenkämper et al., 2017)) and domain-specific tools designed for coral reef analysis (*e.g.*, PhotoQuad (Trygonis and Sini, 2012), CPCe (Kohler and Gill, 2006), CoralNet (Chen et al., 2021b), ReefCloud (2025), and TagLab (Pavoni et al., 2022)).

We summarized four critical features that typically underpin user requirements. To begin with, users need the ability to annotate regions of interest, either through sparse point annotations or dense segmentation masks, to identify objects within those regions. Next, classification of these regions is required to distinguish between background, coral, or others. After processing the image, the tool should generate statistical reports based on the processed visual data to provide statistical results. Finally, the capability to export data is also critical for enabling data sharing, visualization, communication, and further analysis.

While CoralSCOP-LAT provides most of the functionalities supported by existing tools (with the exception of sparse point-based approaches), it also incorporates a deep learning coral reef segmentation model CoralSCOP (Zheng et al., 2024c) to automatically detect dense masks to represent coral regions in target images. This feature significantly enhances the annotation and classification process, providing an efficient and user-friendly solution for coral reef analysis.

### 3.2. Analysis effort efficiency

In this section, we demonstrate that CoralSCOP-LAT is able to provide a more efficient method for coral reef image analysis by reducing the user's operational effort while maintaining critical detail, such as boundaries and small coral areas. We evaluate the method using a pixel-level analysis to compare the effort required to accurately label each pixel in the images. The annotation process is simulated using three approaches: sparse point-based methods (*e.g.,* CPCe, CoralNet), SAM-based methods (*e.g.,* CVAT, TabLab), and CoralSCOP-LAT.

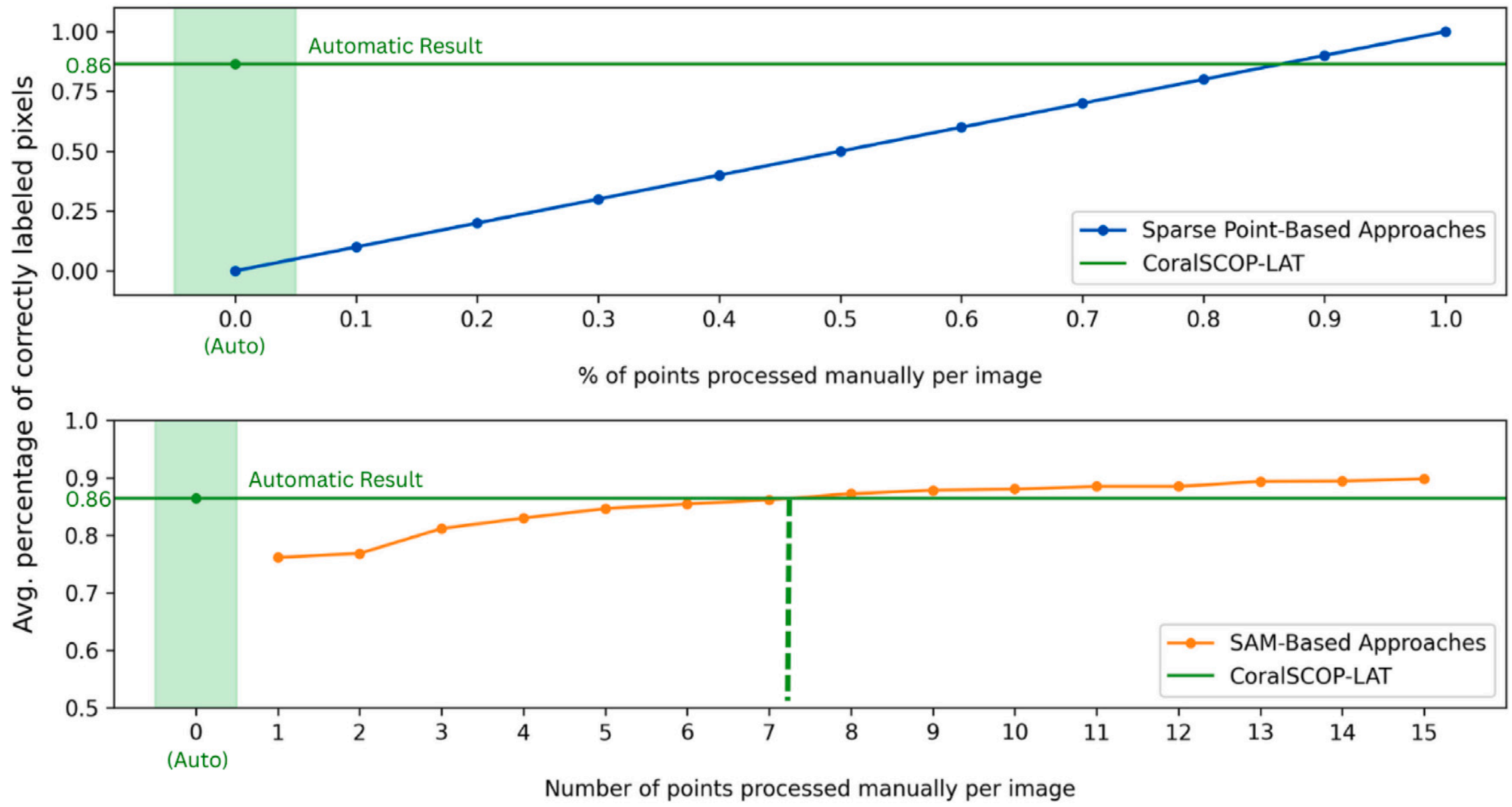

**Fig. 9.** Pixel accuracy versus manual annotation effort for different approaches. CoralSCOP-LAT can automatically replace the effort of annotating seven points in SAM-based approaches, while sparse point-based approaches demand substantial effort for comparable results.

**Dataset.** The evaluation was performed using the CoralMask (Zheng et al., 2024c) testing dataset, which consists of 830 coral reef images along with their corresponding segmentation masks. These masks serve as ground truth, indicating coral reef area.

- CoralSCOP-LAT: To facilitate large-scale comparisons, we directly utilize the unrefined predictions generated by CoralSCOP as the output of CoralSCOP-LAT. This represents the **lower bound** of CoralSCOP-LAT's performance, as no manual refinement of the predictions was performed by the user.
- Sparse point-based approaches: We simulate user behavior using an automated pipeline. For each image, a fixed number of sparse points are randomly scattered, and each point is classified as *coral* or *non-coral* based on the ground truth segmentation mask.
- SAM-based approaches: Similar to the sparse point-based approaches, an automated pipeline was used to simulate user interactions. The process begins by randomly selecting a single *coral* pixel from the ground truth segmentation mask, which is then provided as input to the SAM model (ViT-B). The model generates an initial segmentation mask. The subsequent input point is sampled either from the error region (the area where the resulting segmentation mask differs from the ground truth) or from a pixel belonging to another *coral* instance. This iterative process simulates user refinement of the segmentation.

Using the annotations generated by the aforementioned methods, we estimated the pixel accuracy as the percentage of pixels correctly labeled as either *coral* or *non-coral*. Intersection over Union (IoU) was not employed, as it is not directly compatible with both sparse and dense annotation approaches. As illustrated in Fig. 9, the pixel accuracy of sparse point-based approaches increases linearly with the percentage of labeled pixels, which also indicates that significant manual effort is required to approach segmentation-level accuracy. In contrast, CoralSCOP-LAT achieves performance comparable to SAM-based approaches using 7 sampled points. Importantly, the reported pixel accuracy of CoralSCOP-LAT represents its *minimum* accuracy, as the predictions were not refined with any human intervention. This indicates that the accuracy can be further improved through manual refinement of the segmentation masks. Therefore, CoralSCOP-LAT effectively minimizes the initial annotation effort by providing high-quality segmentations.

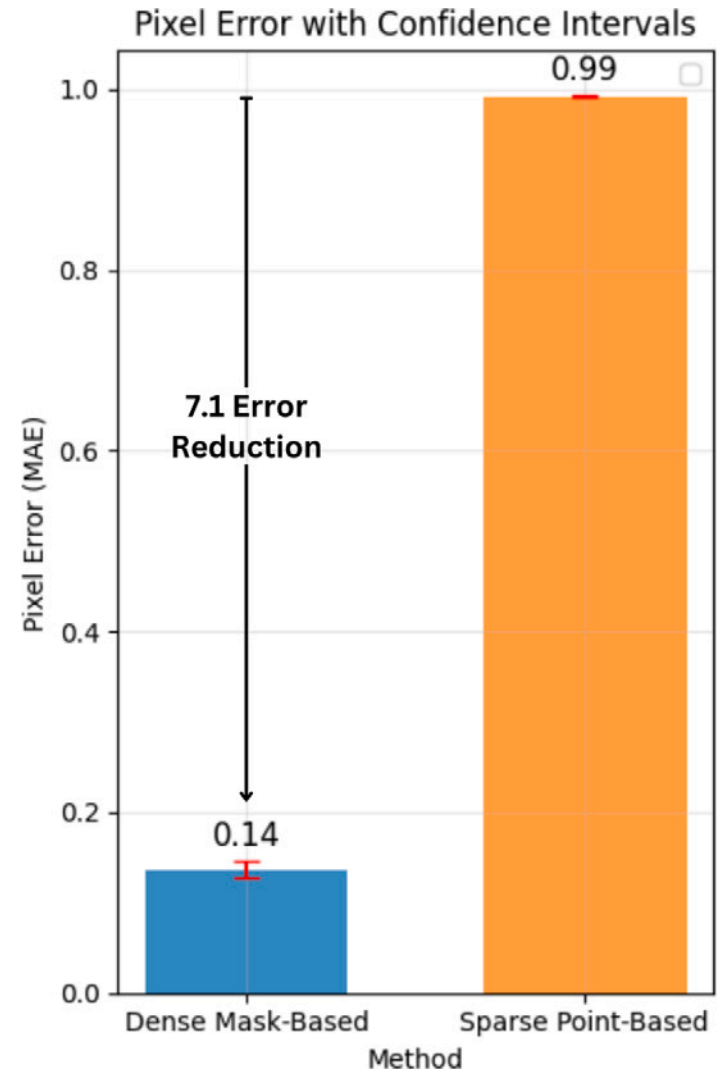

**Fig. 10.** Comparison of annotation error between sparse point-based approaches and dense mask-based approaches. The red whiskers represent the confidence interval (95%). (For interpretation of the references to color in this figure legend, the reader is referred to the web version of this article.)

### 3.3. Annotation accuracy

We contend that sparse point-based approaches are inherently limited in capturing the full features of corals, as they tend to overlook small corals and irregular boundaries. In this section, we demonstrate that using dense segmentation masks can accurately and precisely identify coral regions within images, addressing the shortcomings of sparse-point-based methods such as CPCe (Kohler and Gill, 2006).

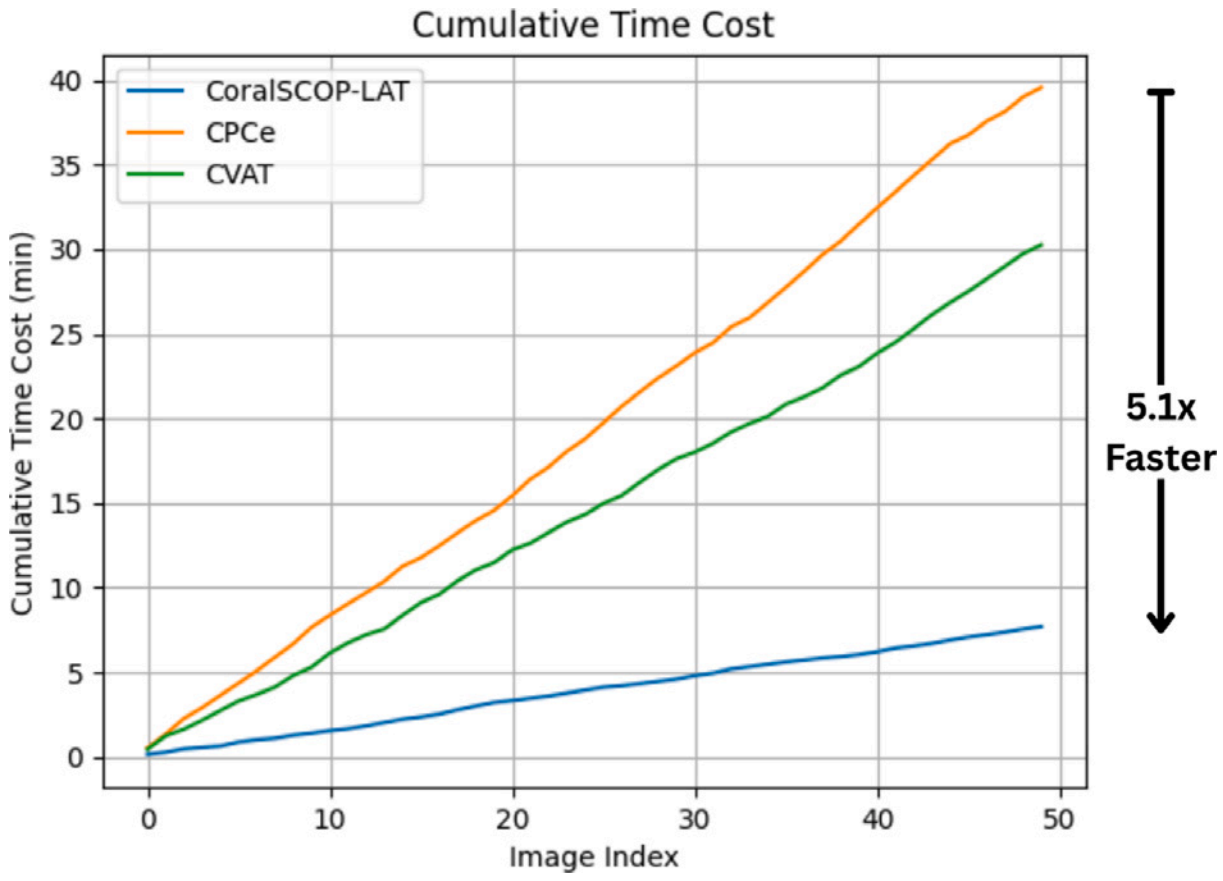

**Fig. 11.** Time cost for using CPCe, CVAT, and `CoralSCOP-LAT`. (For interpretation of the references to color in this figure legend, the reader is referred to the web version of this article.)

We follow the same procedure in Section 3.2, and we take CPCe as the baseline tool for comparison. We generate the prediction from both CPCe and our tool, and then we evaluate the mean absolute error (MAE) after processing the entire CoralMask testing set. The MAE for evaluation is defined as:

$$\mathrm{MAE} = \frac{1}{N}\sum_{i=1}^{N}\left|y_i - \hat{y}_i\right|, \tag{1}$$

where $N$ is the total number of pixels in image, $y_i$ is the ground truth label of the pixel, and $\hat{y}_i$ is the predicted label. We set the number of sampled points to 5000, which is the maximum capacity that CPCe can process without crashing.

As shown in Fig. 10, `CoralSCOP-LAT` achieves a substantial improvement, reducing the cumulative MAE by a factor of 7.4. This improvement is attributed to `CoralSCOP-LAT`'s ability to generate dense segmentation masks, providing predictions for every pixel in the image. In contrast, CPCe only considers a fixed number of the sampled points, which inherently limits its accuracy and precision. This limitation leads to the inevitable under- or over-estimation of coral regions.

### 3.4. Time efficiency

In this section, we evaluate the analysis time cost required for CPCe (sparse-point based approach), CVAT (dense segmentation based approach), and `CoralSCOP-LAT`. Additionally, we assess the impact of integrating an automatic coral segmentation algorithm, CoralSCOP, which is the primary distinction between `CoralSCOP-LAT` and other dense segmentation tools, such as CVAT and TagLab.

To empirically investigate these differences, 15 volunteers were invited to annotate 50 images using the provided tools. The participants were evenly assigned to three groups: those using CPCe (with 100 sampled points per image), those using CVAT, and those using `CoralSCOP-LAT`. The ground truth coral region and corresponding genera are provided as a reference to ensure all candidates have the same knowledge level to perform the analysis. Each image contained an average resolution of 2356 × 1508 pixels and was presented in the same order. We measure the time required for each user to complete the annotation process of each image, and measure the cumulative time consumed.

**Table 2**
Time needed to process 50 images using CPCe, CVAT, and CoralSCOP-LAT.

| Tools | Average time cost (min) | Confidence interval (95%) |
| --- | --- | --- |
| CPCe | 39.6 | [39.2, 40] |
| CVAT | 30.3 | [30, 30.5] |
| CoralSCOP-LAT | 7.7 | [7.5, 7.8] |

The 50-image subset is extracted from the dataset that was captured during a reef check[1,2] survey conducted by Tursiopsdivers[3] on the 10th of August 2024. The survey covered the 100 m transect of the Hong Kong reef check site no. #3 (N22° 32.511 E114° 26.275 N22° 32.467 E114° 26.320, Wong Ye Kwok, Tung Ping Chau, here add reference to AFCD and reef check org). A GoPro Hero9 camera was used for video data collection (1920 × 1080 resolution at 60 fps), capturing the 100 m transect. The images were generated from the video post-survey, capturing all coral colonies along the transect line. A coral health chart card was utilized during data collection.[4] The type of the corals, the health of the corals, and the genus of the coral colonies were identified post-survey.

As illustrated in Fig. 11 and Table 2, `CoralSCOP-LAT` significantly reduces the time cost of the annotation process by 5.1 times. This improvement can be attributed to key differences in the annotation workflow. CPCe requires users to annotate each sampled point in an image, regardless of the number or size of coral reefs present. Consequently, participants must annotate all 100 points in an image, even if only a single coral instance is present. In contrast, dense segmentation tools, such as CVAT and TagLab, allow users to generate dense segmentation masks only when they identify the presence of coral. Segmentation-based approaches therefore eliminate redundant annotation efforts. Due to the intricate structures of coral and the challenges posed by underwater imaging, such as blurriness, color distortion, and light attenuation, users must carefully delineate the boundaries of each coral instance. `CoralSCOP-LAT` further reduces this workload by automatically segmenting coral regions in the images. Users are only required to assign labels to the corresponding segmentation masks and adjust the coral regions in cases where inaccuracies are detected.

### 3.5. Annotation flexibility

Annotation flexibility is crucial for the annotation tools to provide user-aligned analysis, as it reflects the ability to align with user requirements. `CoralSCOP-LAT` and CPCe offer varying levels of user control over annotations.

- CPCe limits user control in point annotations. Once random points are generated, users cannot modify their locations, even if the randomly sampled sparse points do not adequately cover the coral region. Users can only increase the number of points or regenerate all the sparse points, without the ability to modify or specify their locations, restricting the ability of CPCe to meet user expectations.
- `CoralSCOP-LAT` offers users full control over mask annotations. Users can freely add or remove masks using the `ADD MASK` and `REMOVE MASK` functions. User inputs are converted into point prompts for the backend model, which then generates the corresponding masks. This adaptability allows for more precise and customized annotations, accommodating diverse coral structures and enhancing the overall user experience.

[1] Hong Kong Agriculture, Fisheries and Conservation Department: https://www.afcd.gov.hk/english/conservation/con_mar/con_mar_cor/con_mar_cor_hkrc/con_mar_cor_hkrc5.html.

[2] Reef Check Webpage: https://www.reefcheck.org/.

[3] Tursiopsidvers Webpage: https://www.tursiopsdivers.com/.

[4] Coral Watch Program: https://coralwatch.org.

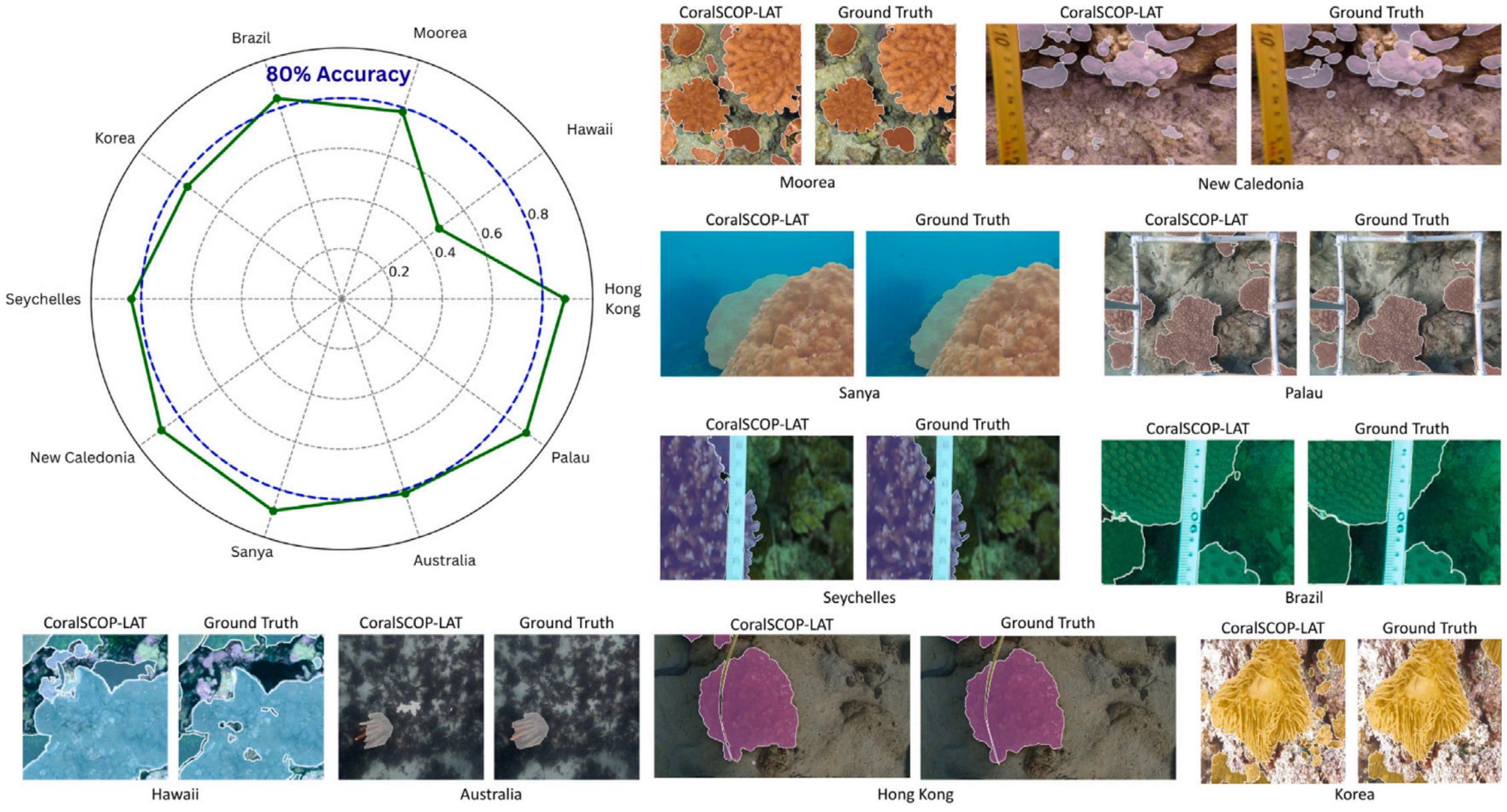

**Fig. 12.** CoralSCOP-LAT demonstrates strong generalization capability, achieving around 80% pixel accuracy in the majority of evaluated locations. Moorea images are from Under The Pole.

**Table 3**
Average pixel accuracy for each location.

| Location | Avg. pixel accuracy | Location | Avg. pixel accuracy |
|---|---|---|---|
| Brazil | 0.8373 | Moorea | 0.7838 |
| Hawaii | 0.4778 | New Caledonia | 0.8661 |
| Hong Kong | 0.8888 | Palau | 0.9061 |
| Sanya | 0.8887 | Australia | 0.8165 |
| Korea | 0.7622 | Seychelles | 0.8411 |

### *3.6. Generalization ability*

In this section, we evaluate the generalization capability of CoralSCOP-LAT by assessing the pixel accuracy across ten geographically diverse locations: Brazil, Korea, Hawaii, Hong Kong, Sanya, Moorea, New Caledonia, Palau, Australia, and Seychelles. Each location comprises 75 images accompanied by corresponding ground truth segmentation masks. Consistent with the methodology employed in prior experiments, the results produced by CoralSCOP-LAT are entirely automated and have not undergone any human refinement.

The source imagery utilized in this analysis originates from previously published work: Brazil (Suggett et al., 2012), New Caledonia (Camp et al., 2017), Seychelles (Gardner et al., 2019), Hong Kong (Ziqiang et al., 2024), Sanya (Jiang et al., 2025), Australia (Bewley et al., 2015) and Hawaii (Edwards et al., 2017). Imagery for the remaining locations was provided through collaborations with biologists and divers.

The experimental results, along with quantitative visualizations, are presented in Fig. 12 and Table 3. While the performance of CoralSCOP-LAT varies across different locations, it managed to maintain over 80% pixel accuracy across most of the locations. However, a noticeable decline in performance is observed for Hawaii due to the presence of false-negative pixels, where CoralSCOP-LAT has insufficient confidence in classifying certain pixels as *coral*.

## 4. Discussions

CoralSCOP-LAT reformulates the workflow for coral reef analysis and bridges the gap between coral ecologists and state-of-the-art coral reef segmentation models. We detail our discussion as follows:

- **Sparse vs. Dense**: CoralSCOP-LAT emphasizes the advantages of utilizing dense segmentation masks for representing the *coverage* and *population* of coral reefs. Existing sparse point-based analytical approaches are constrained by their inevitable downsampling nature during the point sampling procedure and suffer from the limited ability to capture detailed morphological structures and intricate coral boundaries. In contrast, dense segmentation masks provide pixel-level annotations, which enable a more comprehensive representation of coral morphology and ensure better detection of small coral colonies that point-based analytical approaches might overlook. By utilizing dense segmentation masks as the analysis format, CoralSCOP-LAT enhances both the precision and reliability of coral reef analysis.
- **Accessibility to advanced models**: Recognizing the critical role of dense segmentation mask annotation, CoralSCOP-LAT offers a direct and interactive interface to generate coral segments, in an automatic or semi-automatic manner. Recent advances in powerful foundation models (*e.g.*, SAM and CoralSCOP) have drastically reduced the labor intensity associated with generating dense segmentation annotations. However, the difficulty associated with deploying and configuring these models hinders widespread usage for domain research, particularly among coral ecologists with limited technical expertise. CoralSCOP-LAT addresses this barrier by providing an accessible interface to these sophisticated models, thereby broadening their usability and ensuring wider adoption in the coral reef community for scalable and efficient coral reef analysis.
- **Tailored designs for coral ecology**: General-purpose annotation tools often lack tailored features to fulfill domain-specific requirements posed by coral ecology. In detail, existing tools (*e.g.*,

CVAT, BIIGLE) prioritize generalizability across diverse applications. However, these tools do not take the unique demands of coral reef studies into consideration. Thus, coral ecologists must invest additional non-ignorable efforts in adapting the analysis outputs of such tools to derive ecological reports. In contrast, CoralSCOP-LAT is designed to align closely with domain-specific requirements by incorporating features such as *coral health assessments*, *statistical report generation*, and *customizable coral categories*. More importantly, the proposed CoralSCOP-LAT could support automatically generating coral reef masks to significantly reduce human efforts to generate coral cover statistics. These tailored functionalities establish CoralSCOP-LAT as a more effective and suitable analytical tool for coral ecologists.

## 5. Limitation

While CoralSCOP-LAT facilitates automatic coral segmentation using the advanced coral reef foundation model, it is not without limitations. These limitations can be categorized from both technical and biological perspectives:

1. **Technical limitation**: The integrated foundation model inherently requires additional computational resources. Specifically, CoralSCOP-LAT demands 5.7 GB of GPU memory for optimal performance, which restricts its use in some edge devices.
2. **Biological limitation**: CoralSCOP-LAT does not include functionality for automatic coral species classification due to the taxonomic instability[5] for coral species. The definitions of coral species are subject to change over time, making it challenging to establish a stable taxonomic tree that encompasses all coral species. Re-training the classification model with ongoing taxonomic updates alongside the segmentation model is impractical. Consequently, we decided to allow users to define semantic labels based on their own site-specific requirements.

## 6. Border implication

CoralSCOP-LAT makes a broad contribution to the field of coral ecology by providing advanced mask-based coral analysis tools and methodologies to improve the study and monitoring of coral reefs. Its applications span multiple critical areas:

1. **Coral reef surveying**: The pixel-level annotations generated by CoralSCOP-LAT enable precise measurement of key environmental indices, such as coral coverage and bleaching percentages. By facilitating a more accurate assessment of coral reef health, the proposed tool contributes to the early identification of ecological stressors, including coral bleaching events and the spread of invasive species, thereby supporting preventative conservation strategies.
2. **Temporal growth monitoring**: CoralSCOP-LAT is designed to process and analyze diverse visual datasets, including photoquadrats captured at fixed locations over time. Through its advanced segmentation techniques and coral ID labeling, researchers can efficiently track temporal changes in coral structures, such as growth rates and morphological changes, offering valuable insights into reef dynamics.
3. **Video-based analysis**: CoralSCOP-LAT allows users to annotate keyframes from transect videos and integrate these annotations with video segmentation models like SAM 2. The video segmentation models could facilitate the estimation of coral coverage across large spatial scales. This integration could effectively enhance the efficiency and accuracy of video-based coral reef assessments, enabling comprehensive ecological monitoring.
4. **Advancing biodiversity analysis**: CoralSCOP-LAT provides an efficient methodology for annotating irregular objects, and its utility need not be limited to coral ecosystems. The framework can be extended to annotate algae, seagrasses, or other marine organisms, enabling comprehensive statistical analyses and environmental surveys (Voskanyan et al., 2024; Nelli et al., 2024).
5. **Promoting citizen science**: One of the key advantages of CoralSCOP-LAT is its accessibility, as it provides non-specialist users with straightforward access to advanced machine learning models. This feature makes it a valuable tool for citizen science initiatives. Citizen scientists could utilize the framework to explore marine environments and contribute to large-scale datasets by annotating marine species accurately. Such contributions could enhance community involvement in environmental monitoring and education while fostering collaborative research (Voskanyan et al., 2024; Kelly et al., 2020).

## 7. Conclusion

This work introduces CoralSCOP-LAT, the first dense segmentation-based tool for automated coral reef labeling and analysis. By leveraging automation in coral reef segmentation, CoralSCOP-LAT significantly reduces the need for intensive human effort and expert involvement in large-scale coral reef monitoring and analysis. The tool features hierarchical and versatile functionalities, enabling users to validate segmentation masks, assign semantics, add new annotations, monitor coral status, visualize statistical results, and export data efficiently.

Key contributions of this work include:

- Integration of a state-of-the-art foundation model for robust zero-shot coral detection and segmentation.
- Development of hierarchical features and an intuitive interface for streamlined annotation and analysis.
- Achieving higher accuracy and faster performance compared to existing tools.

CoralSCOP-LAT emphasizes the importance of the use of dense masks as the representation of coral reefs and serves as a bridge between coral ecologists and cutting-edge coral segmentation models.

## CRediT authorship contribution statement

**Yuk Kwan Wong:** Writing – review & editing, Writing – original draft, Visualization, Validation, Software, Methodology. **Ziqiang Zheng:** Writing – review & editing, Methodology, Data curation. **Mingzhe Zhang:** Validation, Methodology. **David J. Suggett:** Writing – review & editing, Validation, Supervision. **Sai-Kit Yeung:** Writing – review & editing, Supervision, Project administration.

## Declaration of competing interest

The authors declare that they have no known competing financial interests or personal relationships that could have appeared to influence the work reported in this paper.

## Acknowledgments

We first thank all the data provider:

Section 3.4: We thank Tursiopsdivers for sharing their data in Section 3.4, allowing us to test our software. A special thanks to Dr Eszter Matrai for providing all manual coding for coral health, type, and ID.

Section 3.6: Data presented in Section 3.6 is drawn from previously published works as described in the main text via works led by David Suggett with Emma Camp University of Technology Sydney; David Smith, MARS Sustainability; and/or, Riccardo Rodolfo-Metalpa, Institut de Recherche pour le Développement. We gratefully acknowledge their contribution to this data. For Moorea and French Polynesia photo-quadrats (pictures), we thank Under The Pole, particularly Emmanuelle Périe-Bardout and Ghislain Bardout, and also Dr. Gonzalo Perez-Rosales, for sharing the images and his continuous biological coral analysis support. The image collection was funded by the ANR DEEPHOPE (ANRAAPG 2017, Grant No. 168722), the Under The Pole III Expedition, the Délégation à la Recherche DEEPCORAL, the CNRS DEEPREEF, the EPHE and IFRECOR. No export permits are required for photo-quadrats (pictures) in French Polynesia. We thank Eveline van der Steeg for providing the Palau data. The photos were captured around Ngerchong Island on 4 December 2018 with a TG-6 camera in Underwater Wide mode, with white balance adjusted. We also thanks Professor Jeong Ha Kim for providing the Korea data.

[5] https://www.coralsoftheworld.org/page/overview-of-coral-taxonomy/.

Additionally, we thank Pal's Ltd. for their design of the user interface for the analysis tool.

The work was partially supported by the Marine Conservation Enhancement Fund (MCEF20107 and MCEF22112), an internal grant from HKUST (R9429), HKUST Marine Robotics and Blue Economy Technology Grant . The authors would like to express their sincere gratitude to the "Sustainable Smart Campus as a Living Lab" (SSC) program at HKUST for its invaluable support. This program and staff not only provided essential funding and coordination, but also integrates sustainability into campus operations, which served as the living demonstration of the principles underpinning this research.

## Data availability

The implementation code of `CoralSCOP-LAT` is available on GitHub.

The data used in the experiment are available at Zenodo.